\documentclass[letterpaper, 10 pt, conference]{ieeeconf}

\IEEEoverridecommandlockouts                              

\usepackage[utf8]{inputenc}
\usepackage[T1]{fontenc}

\title{ \LARGE \textbf{Developing cooperative policies for multi-stage tasks}}

\author{Jordan Erskine$^{*}$, Chris Lehnert
\thanks{$^{*}$J. Erskine and C. Lehnert are with the Queensland University of Technology (QUT), Brisbane, Australia {\tt\small \{jl.erskine,c.lehnert\}@qut.edu.au}}
}

\usepackage{comment}
\usepackage{amsmath}
\usepackage{amssymb}
\usepackage{graphicx}
\usepackage{float}
\usepackage[ruled, noline]{algorithm2e}
\begin{document}
\maketitle
\thispagestyle{empty}
\pagestyle{empty}


\begin{abstract}
    \textbf{This paper proposes the Cooperative Soft Actor Critic (CSAC) method of enabling consecutive reinforcement learning agents to cooperatively solve a long time horizon multi-stage task. This method is achieved by modifying the policy of each agent to maximise both the current and next agent's critic. Cooperatively maximising each agent's critic allows each agent to take actions that are beneficial for its task as well as subsequent tasks. Using this method in a multi-room maze domain, the cooperative policies were able to outperform both uncooperative policies as well as a single agent trained across the entire domain. CSAC achieved a success rate of at least 20\% higher than the uncooperative policies, and converged on a solution at least 4 times faster than the single agent.}
\end{abstract}

\section{Introduction}
Reinforcement learning (RL) is a state of the art technique for developing solutions to tasks which are typically challenging to hard code solutions. However, RL methods struggle in high dimensional complex domains. Exploring the domain sufficiently to discover an effective strategy without expending exhaustive amounts of computational resources can be a significant challenge. Real world tasks have the limitation that exploration has to be done real time, far slower than running in simulation. Most robotic tasks need to be learned in the real world, as most simulators are not an accurate representation of the real world. For these reasons, the success of RL approaches to robotics tasks are limited \cite{kober2013reinforcement}. To assist in solving these issues, commonly the domain will be simplified significantly (i.e. discretizing the state-action space). One such approach involves splitting the domain into smaller sub-problems that can be solved individually and then used together to solve the more complex original problem \cite{krishnan2016hirl}. 

To decompose a reinforcement learning task into smaller subproblems, each subproblem has to be defined as its own reinforcement learning task, including an appropriate reward signal. A consequence of splitting up the problem is that agents are solving independent reward signals and the incentive to complete the overall task is lost, likely resulting in a failure to solve the overall task effectively or even at all. This problem is caused by the greedy nature of individual agents only accounting for their subtasks. Without any incentive to cooperate with the next agent, each agent will likely produce behaviours that are only beneficial for their own task, reducing the likelihood of the next agent successfully completing its task. 

\begin{figure}
    \centering
    \includegraphics[width = \linewidth]{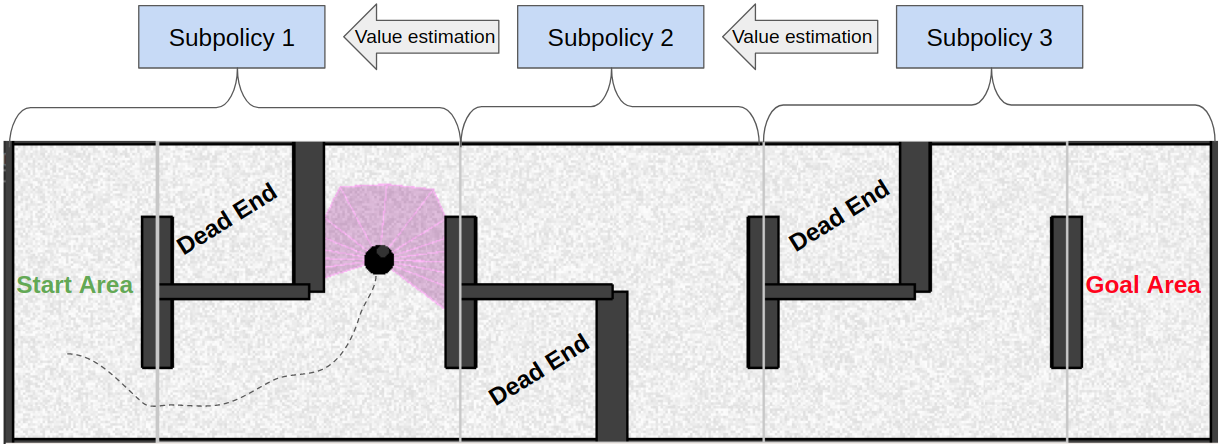}
    \caption{A high level view of the Cooperative Soft Actor Critic algorithm. Each subpolicy acts within its own subtask, and is informed by the next subpolicy about how to act to assist in completing the subsequent task. In the maze domain, the next subpolicy can give information about where the dead ends are and how to avoid them. }
    \label{fig:overview}
\end{figure}

The solution proposed in this paper addresses this disconnect between subtask solution and overall task solution. In our proposed approach each subagent is incentivised to cooperate with the next subagent by training the subagent's policy network to produce actions that maximise the current subagent's critic as well as the next subagent's critic (as shown in Figure \ref{fig:overview}). By incorporating the next subagent's critic, the current subagent can continue to solve it's own subtask while also producing a solution that is beneficial for the next subtask. The main contributions of this paper are:
\begin{itemize}
    \item A novel cooperative policy method, the Cooperative Soft Actor Critic (CSAC) algorithm, which enables agents to learn behaviours that maximise reward for their own task while accommodating for subsequent tasks improving performance for learning long complex sequences of tasks.
    \item A case study in a continuous state/action space maze domain demonstrating CSAC outperforming both an uncooperative policy and single end-to-end policy trained on the same task. In these experiments, CSAC had a success rate at least 20\% higher than the uncooperative policies, and converged on a solution at least 4 times faster than the single policy.
\end{itemize}

\section{Related Work}

Reinforcement learning has been used to solve many different tasks, including exceedingly complex domains, such as the games of Go \cite{Silver2017} and DOTA \cite{OpenAI_dota}. Conventional RL approaches involve learning in an end-to-end manner, training to maximise a reward signal produced by the environment. Although some of the applications are impressive, they are sample inefficient when training \cite{chua2018deep}, sometimes taking billions of training steps in simulation. If RL methods are to be applied to more complex real world scenarios, such as learning robotic manipulation tasks, different methods need to be employed that are able to significantly increase the sample efficiency.

Hierarchical reinforcement learning (HRL) has been a long standing topic in RL \cite{precup1998theoretical} \cite{nachum2018data} \cite{vezhnevets2017feudal} \cite{precup2001temporal}. Hierarchical RL methods seek to capitalise on some of the inherent structure that is present in many tasks. This is commonly done by developing a series of temporally extended macro-actions to solve a task \cite{botvinick2012hierarchical}. These macro-actions are essentially a series of actions that produce a specific outcome, and as such, they are often treated as subtasks if they are undefined and their behaviour needs to be learned \cite{bacon2017option}. The main benefit of using HRL is that solving a series of smaller, simpler subtasks is easier and faster to learn than a single, more complex task \cite{frans2017meta}. This is caused by the fact that in larger, more complex domains, the critic function needs to accurately produce values across more of the state space, a state-space that could require disjointed value approximations \cite{sutton2018reinforcement}. This means that random batches will be less likely to be representative of the state space, and so more batches are needed to refine the function. Using smaller subtasks circumvents this issue.

Many hierarchical reinforcement learning approaches have been designed to deal with sequential tasks \cite{oh2017zero} \cite{icarte2018using}. Typically they have the form of using a meta-controller in conjunction with a series of lower-level controllers for each subtask \cite{sutton1999between} \cite{konidaris2012transfer}. The use of a meta-controller allows for more generalisation, as the series of subtasks can be combined in more versatile ways. This does require more stable and refined sub-policies, and it does not take advantage of the fundamental order that some hierarchical tasks have. These methods follow one of two approaches; manual definition of each subtask, or learning of the subtasks. 

Learning subtasks is an approach to HRL that seeks to decompose the task into subtasks, as well as learn the solutions to these subtasks, autonomously. There have been many tasks solved using this approach \cite{konidaris2012transfer} \cite{Andreas2017} \cite{haarnoja2018latent}. This approach ensures that the subtasks are useful for the overall task, but this tends to be very slow to train. 

Manually defining subtasks is an alternative approach to HRL. This approach involves decomposing a task into subtasks manually, defining the reward signal and termination signal for each subtask \cite{Andreas2017} \cite{forestier2017intrinsically}. This process has two main limitations; the utility of the reward signal, and the combinability of the resulting subtasks. As the reward needs to be manually defined, the behaviours that are incentivised to be learned are ones that the expert deems as correct, which can limit the full potential of the subtask solution. The issue of combinability comes from the fact that the subtasks under this approach are separate from the overall task. The subtasks are incentivised to solve their own task, with no requirement to use a solution that assists in the solution of the overall task. This means that manually defined subtasks can be prone to poor overall task completion. This can be mitigated by further refinement of the reward signal, however, this further biases the solution space and can be an expensive process.

The current literature in the multi-stage reinforcement learning domain have a few main limitations. Firstly, end-to-end learning is an expensive process that scales exponentially in difficulty as the complexity of the task increases. Secondly, most hierarchical approaches do not deal with sequential tasks, instead trying to solve separate subtasks individually, which disconnects that subtasks from the overall task. Those implementations that do deal with sequential tasks do not integrate the sequential nature of their tasks into account when developing their solution. Thirdly, methods that attempt to decompose tasks into subtasks through learnt methods suffer from even higher levels of sample inefficiency than end-to-end learning. The cooperative policies method introduced in this paper utilises a task that has manually been decomposed into subtasks, and then solves the task efficiently by taking advantage of the sequential nature of the task. 

\section{Background}

\subsection{Reinforcement Learning}

We consider the standard Markov Decision Process formulation of solving the problem of picking optimal actions to solve a task. At each timestep \(t\) an agent can take an action \(a_t\) from the current state \(s_t\), which results in the environment evolving to the next state \(s_{t+1}\), and the agent receives a reward \(r_t\). The RL process involves an agent learning a policy \(\pi\) that can produce actions that maximise the return from the environment. The return in most RL approaches use a discounted sum of future rewards, given by the following:

\begin{equation}
    R = \sum^{\infty}_{t=0}\gamma^{t}r_t
\end{equation}

Where \(\gamma\) is a discount factor for future rewards. Many RL algorithms utilise an estimate of the expected return from a state with a specified action, often referred to as a Q function, or critic.

\begin{equation}
    Q(s_t,a_t) = \mathbb{E}[R_t|s_t=s,a_t=a]
\end{equation}

Utilising the fact that the discounted sum of future rewards can be recursively defined, a self-consistent equation can be constructed, commonly known as the Bellman equation, as shown below.

\begin{equation}
    Q(s_{t+1},a^*_{t+1}) = r_t+\gamma Q(s_t,a^*_t)
\end{equation}

Where \(a^*_t\) is the optimal action given the state \(s_t\). Using the Bellman equation the critic function can be refined. Using this critic function, a policy function can be trained to produce actions that maximise the expected return.

\subsection{Soft Actor Critic}

The CSAC algorithm is an extension of the Soft Actor Critic (SAC) algorithm \cite{haarnoja2018soft}. It utilises a Q value function \(Q_\beta(s,a)\), known as a critic, approximated using a neural network with parameters \(\beta\) and a policy \(\pi_\psi(s)\) approximated using another neural network with parameters \(\psi\). SAC is an off-policy algorithm, so typically each transition of state, action, reward, and next state (\(s,a,r,s'\)) are stored in a replay buffer B.

The critic is trained by back-propagating the gradient from a loss function defined as the mean squared Bellman error, derived from the Bellman equation:

\begin{equation}
    Q_{loss}=\frac{1}{M}\sum^{M}_{n=0}(Q_\beta(s_n,a_n)-(r_n+\gamma Q_\beta(s'_n,\pi_\psi(s'_n))))^2
\end{equation}

Where M is the sample size of a batch of transitions sampled from B, with \(n\) being the index of a sample from the batch. The policy is trained by maximising the value of the critic.

The SAC algorithm uses a few techniques to improve the stability of the learning process. Firstly, it introduces a second critic and uses the minimum value between the two when updating the policy. This smooths the function approximation and addresses some of the overestimation bias issues that affect many value gradient RL algorithms. The SAC algorithm also adjusts the policy network so that rather than outputting a deterministic action, it instead outputs a Gaussian distribution from which an action is sampled. This network is then trained to maximise entropy as well as the future expected reward, according to the following equation:

\begin{equation}
    \pi_{loss}=\frac{1}{M}\sum^{M}_{n=0}(\alpha \log(\pi_\psi(a|s))-Q_\beta(s',\pi_\psi(s')))
\end{equation}

Where \(\alpha\) is a variable designed to tune the exploration bias. This adjustment encourages the policy to explore where possible while exploiting its ability to complete the task.

\begin{figure*}
    \centering
    \includegraphics[width=\linewidth]{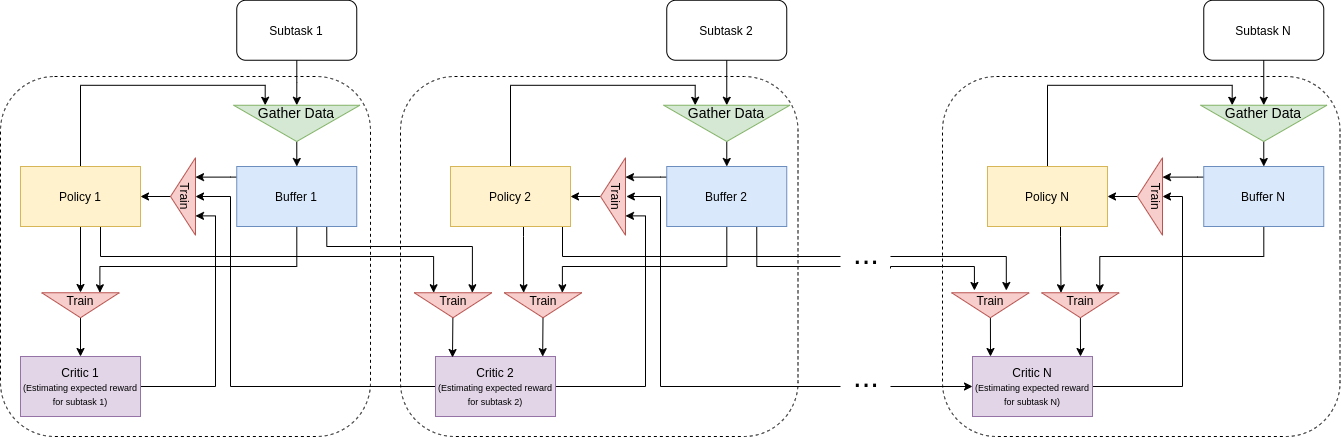}
    \caption{The structure of a cooperative policy implementation in a N subtask environment. Each subpolicy is trained using the current critic and the subsequent critic, though the last subpolicy is only trained relative to its own critic. Each critic is trained using data from the current and previous subtasks, utilising the subpolicy from that subtask.}
    \label{fig:CoopPolDiagram}
\end{figure*}

\section{Method}

This paper introduces the Cooperative Soft Actor Critic (CSAC) algorithm, in which consecutive policies act to cooperate with the subsequent policies. For a policy to be cooperative with another policy, it needs to have a concept of what the other policy wants to achieve. In the traditional Q learning approach to reinforcement learning, the critic is used for this role. The critic informs the policy of how good certain actions and states are to complete its task effectively. This means that a policy can act cooperatively with a subsequent policy by acting in a way that maximises the subsequent critic as well as the current critic.

This method of utilising cooperative policies is done by altering the policy training objective. Instead of training a policy to maximise its critic, this method trains the policy to maximise a convex combination of the current and subsequent agent's critics. This altered architecture is shown in Figure \ref{fig:CoopPolDiagram}. Several changes need to be made to the standard SAC architecture to enable cooperative policies to be applied:
\begin{itemize}
    \item Define the task as a series of \(n\) subtasks, with their own reward signal and a signal that the subtask has been completed. It is assumed that these subtasks are always completed in the same order, and as such, the next subtask is always the same.
    \item Create an agent for each subtask, including a critic, a policy, and a replay buffer that stores the data collected within that subtask.
    \item Collect data in the task, swapping to the next agent when the current subtask is completed, storing all transitions from a subtask within the corresponding replay buffer.
    \item Update each policy according to equations \ref{eq:policyUpdate}, \ref{eq:convComb} and \ref{eq:normalise} using data sampled by the corresponding agent. 
    \item Train each critic on data sampled from the current and previous subtasks. This must be done using the policy that was trained in that subtask.
\end{itemize}

\begin{algorithm}
\caption{Gathering Data}
\label{alg:gather}
\SetAlgoLined
    Environment with N subtasks and associated reward signals \(r_{(1, ..., N)}\)\;
    For each subtask initialise an agent \(a_n\), including a policy \(\pi_n\) with parameters \(\psi_n\), a critic \(Q_n\) with parameters \(\beta_n\), and a replay buffer \(B_n\)\;
    \While{timestep \(<\) maxTimestep}{
        s, n \(\xleftarrow[]{}\) reset environment\;
        \While{not done}{
            a \(\sim \pi_n(s)\)\;
            s', n', \(r_{(1,...,N)}\), done \(\xleftarrow[]{}\) environment step with a\;
            record (s, a, \(r_{1,...,N}\), s', done) in \(B_n\)\;
            s \(\xleftarrow[]{}\) s'\;
            n \(\xleftarrow[]{}\) n'\;
        }
        
    }

\end{algorithm}

\begin{algorithm}
\caption{Cooperative Training}
\label{alg:train}
\SetAlgoLined
    Set of N agents \(a_n\), each with policy \(\pi_n\) with parameters \(\psi_n\), a critic \(Q_n\) with parameters \(\beta_n\), and a replay buffer \(B_n\)\;
    cooperative ratio \(\eta\)\;
    discount factor \(\gamma\)\;
    entropy maximisation term \(\alpha\)\;
    \For{n in (1,...,N)}{
        sample minibatch \(b\) from \(B_n\) of M samples (s, a, \(r_{(1,...,N)}\), s', d)\;
        \textit{train critics}\;
        \For{j in (n, n+1)}{
            \textit{calculate targets}\;
            \(y(r,s',d) = r + (1-\gamma)(Q_j(s',a') - \alpha\) log(\(\pi_n(a'|s')))\), \(a'\sim \pi_n(s')\)\;
            \textit{update critic using one step of gradient descent by applying}\;
            \(\nabla_{\beta_j}\frac{1}{M}\sum(Q_j(s,a) - y(r_j,s',d))^2\)\;
        }
        \textit{train policy}\;
        \textit{generate new actions}\;
        \(a' \sim \pi_n(s)\)\;
        \For{j in (n, n+1)}{
            \textit{normalise critic values across minibatch}\;
            \(Q'_j(s,a') = \frac{Q_j(s,a')-\min(Q_j(s,a'))}{\max(Q_j(s,a')-\min(Q_j(s,a')}\)
        }
        \textit{update policy using one step of gradient descent by applying}\;
        \(\nabla_{\psi_n}\frac{1}{M}\sum(\alpha log(\pi_n(a'|s))-(\eta Q'_n(s,a')+(1-\eta)Q'_{n+1}(s,a')))\)\;
    }
\end{algorithm}

The policy update is modified from the original SAC policy update. Equation \ref{eq:policyUpdate} shows the modified policy update for a policy \(\pi\):

\begin{equation} \label{eq:policyUpdate}
    \pi_{i,loss} = \frac{1}{M}\sum^{M}_{0}(\alpha \log\pi_\theta(a|s) - C(Q_i,Q_{i+1}))
\end{equation}

Where \(M\) is the number of samples in a batch sampled from replay buffer \(B\), \(\alpha\) is the entropy maximisation term, and \(C\) is a convex combination of the current and subsequent critics, as defined by:
\begin{equation} \label{eq:convComb}
    C(Q_i,Q_{i+1}) = \eta Q'_i(s')+(1-\eta) Q'_{i+1}(s')
\end{equation}

Where \(\eta\) is a weighting factor which shall be referred to as the cooperative ratio. This ratio affects how much the current policy acts with respect to the subsequent critic, and is a number between 0 and 1. A cooperative ratio closer to 1 is more incentivised to maximise the current critic's value, whereas a cooperative ratio closer to 0 is more incentivised to maximise the subsequent critic's value. Each Q function is normalised across the batch as denoted by \(Q'\). This normalisation is evaluated using:

\begin{equation} \label{eq:normalise}
    Q'(s) = \frac{Q(s,\pi(s))-\min_{s'\in B}(Q(s',\pi(s')))}{\max_{s'\in B}(Q(s',\pi(s')))-\min_{s'\in B}(Q(s',\pi(s')))}
\end{equation}

This normalisation is done for each critic separately and is recalculated for each batch. This normalisation squeezes the critics output to values between 0 and 1. This normalisation is required to compare the current and subsequent critics, which normally produce arbitrary ranges of values. The full algorithmic approach is shown in Algorithms \ref{alg:gather} and \ref{alg:train}. First data is gathered according to Algorithm \ref{alg:gather}, and then the agents are trained according to Algorithm \ref{alg:train}.

Within the proposed method a cooperative policy requires a subsequent policy to be cooperative with, which the last policy does not have. The final policy is only trained to maximise its own critic. This means that in a task decomposed into \(n\) subtasks, there will be \(n-1\) cooperative policies.

\subsection{Implementation details}

The implementation used in this paper was based on the RLkit implementation of SAC\footnote{https://github.com/vitchyr/rlkit}. This implementation utilises an epoch based approach. Each epoch is composed of a set of evaluation runs, a set of exploration runs, and then a set of training loops. This approach was used as it allowed the exploration environment to be different from the evaluation environment. In this paper, this was utilised to improve exploration across the domain. This was done in the following way; the evaluation domain initialised the agent only within the starting area, whereas the exploration domain initialised the agent in a random location with the entire domain. This allowed the agent to gather experience from across the entire domain, whereas the evaluation only measured the results of completing the entire maze. Both the evaluation and the exploration was done according to algorithm \ref{alg:gather}, though the evaluation data was not stored in the replay buffer. This was then followed by the training loop, which was composed of numerous iterations of algorithm \ref{alg:train}.

A difference between the listed algorithms \ref{alg:gather} and \ref{alg:train} and the actual implementation is that twin critics were used instead of single critics, and each of these utilised a target network. This method is the same as is used in traditional SAC \cite{haarnoja2018soft2}, but was excluded from the algorithms for clarity.

\section{Experimental studies - Maze Domain}


\begin{figure}
    \centering
    \includegraphics[width = \linewidth]{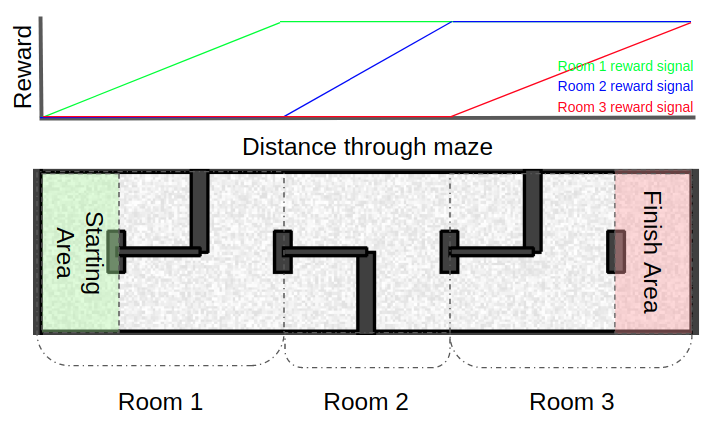}
    \caption{The Maze domain. The agent begins in the starting area and the first subpolicy produces actions to navigate the first room. As the agent enters a new room, the corresponding subpolicy takes charge. The reward signal used in this domain is shown above the maze.}
    \label{fig:MazeDiag}
\end{figure}

In this domain, a series of consecutive rooms were created. Each room had two doors to exit into the next room. Each room was considered a subtask with the goal being to exit the room. The overall task in this domain was to travel through all the rooms to get from one side to the other. The rooms are designed such that an optimal subtask solution would not lead to an optimal overall solution. This domain allows for easy extension in terms of the number of subtasks that are required to be able to solve the overall task. An example of this domain is shown in Figure \ref{fig:MazeDiag}. 

This domain involved each room having a dead-end behind one of the doors, meaning that the fastest way from the current room would lead to a dead-end in the next room. This environment is designed to show the limitations that can arise in the manual definition of reward functions for subtasks, as the greedy approach of solving the current subtask would lead to a dead-end in the next subtask. 

The agent in this domain uses two continuous actions, linear velocity and angular velocity. The agent uses an observation of the environment that includes a laser scan as well as a global position in the maze, all of which are continuous measurements. The implementation uses the hyperparameters shown in Table \ref{tab:Hypers}. These hyperparameters were adapted from the RLkit SAC implementation.

Three different methods were evaluated in this domain; the CSAC algorithm, the uncooperative method, and the single agent method. These three types were selected to highlight the value of using cooperative policies. The CSAC algorithm is the proposed method described in this paper, utilising the critic of the subsequent agent to inform the current agent how to act effectively with regards to future subtasks. The uncooperative method uses a separate agent for each subtask, without any communication between agents. Each agent is attempting to maximise solely its own reward signal. This method represents the naive approach of separating the domain into subproblems. The single agent method involves using a single end-to-end SAC agent that is trained to perform across the whole domain, utilising a reward signal that is the combined reward signal from all the subtasks. This method represents the standard RL approach to solving a task. 

\begin{table}[]
    \begin{center}
    \caption{Hyperparameters used in Maze Domain}
    \label{tab:Hypers}
    \begin{tabular}{||c|c||}
         \hline
         Hyperparameter & Value \\
         \hline \hline
         Gamma (\(\gamma\)) & 0.95 \\
        \hline
        Replay buffer length & 1e6 \\
        \hline
        Batch size & 256 \\
        \hline
        Learning rate & 3e-4 \\
        \hline
        Maximum episode length & 1000 \\
        \hline
        Soft target update factor & 0.005 \\
        \hline
        Evaluation timesteps per epoch & 5000 \\
        \hline
        Exploration timesteps per epoch & 5000 \\
        \hline
        Training loops per epoch & 1000 \\
        \hline
        Epochs & 3000 \\
        \hline
    \end{tabular}
    
    \end{center}
\end{table}

\subsection{Optimal cooperative ratio}

The first study in the Maze domain tested the performance of the three different methods. The CSAC algorithm requires the selection of a cooperative ratio to determine the trade-off between current and subsequent policies. As the cooperative ratio variable cannot be trained, a sweep was conducted across the variable. Three different length mazes were tested using a sweep across the cooperative ratio parameter to determine its effects on learning performance. The best results from this sweep can then be compared to the results from using the uncooperative and single agent methods.

\subsection{Dynamic cooperative ratios}

The previous experiment used a fixed cooperative ratio for each policy for the whole trial. An investigation on whether there is a benefit when using different cooperative ratios across multiple subtasks was conducted. This experiment was conducted in the 3 room maze domain, in which there are two cooperative policies. A parameter sweep was conducted across the cooperative ratio for each cooperative policy to investigate if any trends on learning performance can be found.

\section{Results}

\begin{figure}
    \centering
    \includegraphics[width = \linewidth]{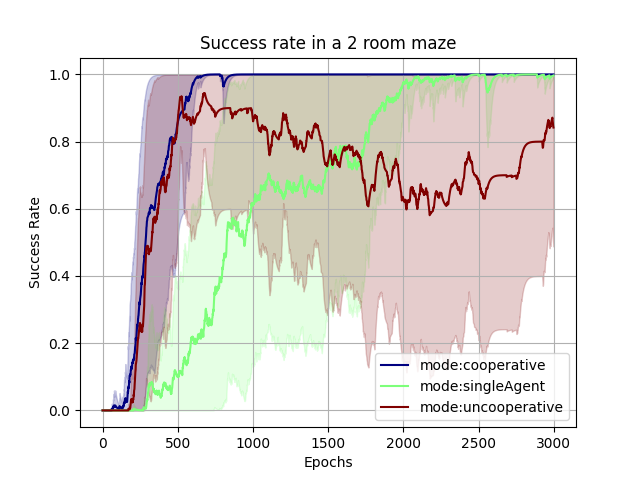}
    \caption{Success rate in the 2 room implementation of the maze domain. The plot shows the average success rate across 10 trials. This uses a cooperative ratio of 0.1}
    \label{fig:2RoomResults}
\end{figure}

\begin{figure}
    \centering
    \includegraphics[width = \linewidth]{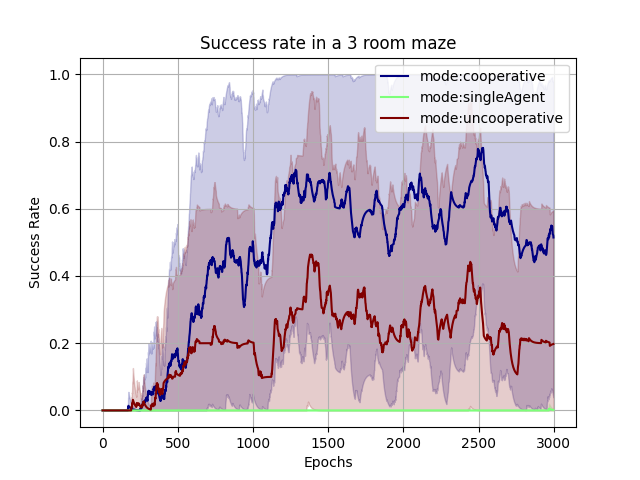}
    \caption{Success rate in the 3 room implementation of the maze domain. The plot shows the average success rate across 10 trials. This uses a cooperative ratio of 0.1}
    \label{fig:3RoomResults}
\end{figure}

\begin{figure}
    \centering
    \includegraphics[width = \linewidth]{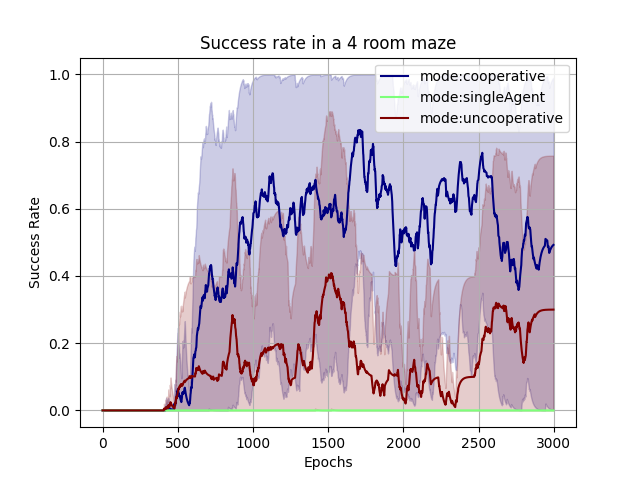}
    \caption{Success rate in the 4 room implementation of the maze domain. The plot shows the average success rate across 10 trials. This uses a cooperative ratio of 0.9}
    \label{fig:4RoomResults}
\end{figure}

    

\subsection{Optimal cooperative ratio}
The results for the 2 room experiment within the maze domain are presented in Figure \ref{fig:2RoomResults}. This experiment shows that the cooperative and uncooperative agents both learn a successful policy in a similar time period, whereas the single agent policy took more than 3 times as long to reach a similar performance. The uncooperative agents, though they reached a high level of performance quickly, had a degenerative solution. This can be explained through the maze design. Each subtask is quite simple to learn, at least to a point where a solution can be produced regularly. As each subtask is further refined, the optimal behaviour for the subtask is discovered, which in this maze domain is to take the dead-end path. This is why performance deteriorates for the uncooperative method, in which each policy is greedy. This experiment shows that decomposing a task into subtasks is beneficial in terms of training speed, shown by the relative training speed of the cooperative and uncooperative policies compared to the single agent experiments. This experiment also shows that just decomposing a task into subtasks and then treating them as entirely separate problems can lead to suboptimal or degenerate solutions. 

Figures \ref{fig:3RoomResults} and \ref{fig:4RoomResults} show the success rates of the three different agent types in the 3 and 4 room mazes. In both of these domains, the single agent was not able to learn a successful policy at all within the 3 million training steps. The cooperative agents were able to achieve a consistently higher performance than the uncooperative policies. This demonstrates the value of splitting a task into subtasks and then solving them cooperatively. 

\begin{figure}
    \centering
    \includegraphics[width = \linewidth]{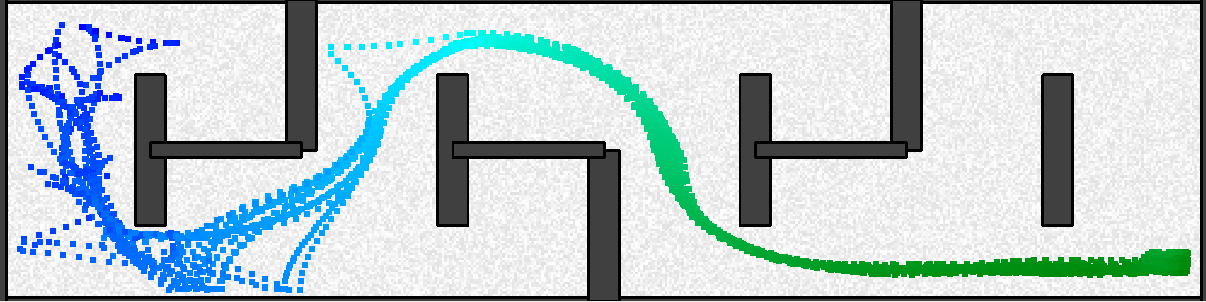}
    \caption{A successful policy in the 3 room domain trained using cooperative policies. This diagram shows 20 runs using different initialisations. The colour coding represents the value estimated by the critic of the second subtask.}
    \label{fig:succRuns}
\end{figure}

\begin{figure}
    \centering
    \includegraphics[width=\linewidth]{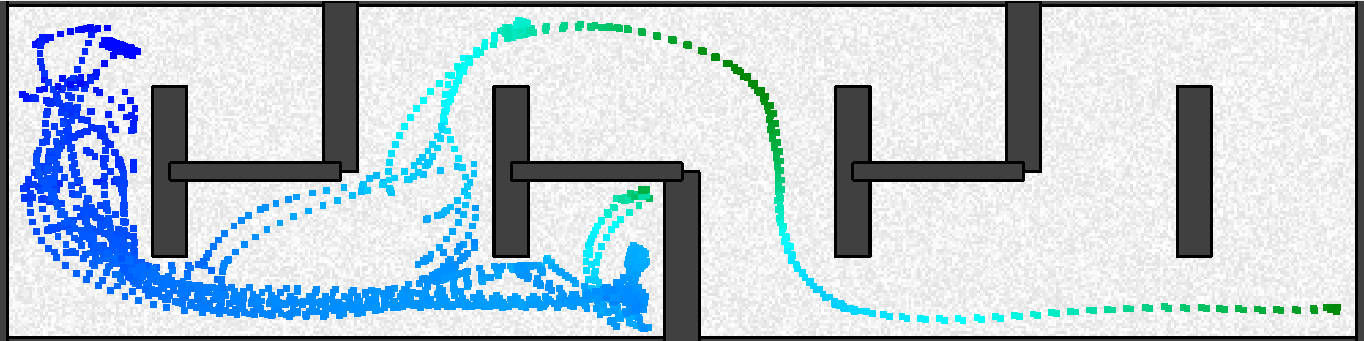}
    \caption{An unsuccessful policy in the 3 room domain trained using uncooperative policies. This diagram shows 20 runs using different initialisations. The colour coding represents the value estimated by the critic of the second subtask.}
    \label{fig:unsuccRuns}
\end{figure}

An example of a successful policy behaviour generated using the CSAC algorithm in the 3 room domain is shown in Figure \ref{fig:succRuns}. When acting cooperatively, each policy is able to avoid the dead ends. An example of a unsuccessful policy generated by the uncooperative method is shown in Figure \ref{fig:unsuccRuns}. These greedy policies are snared by the dead-ends, only avoiding them when it is easier to go around.

Figure \ref{fig:successByCRatio} shows the different success rates in the 3 different mazes with different cooperative ratios. The cooperative ratio is a variable that has a large impact on the performance of cooperative policies. Small changes in this variable can produce very large changes in success rate. The optimal value for the cooperative ratio is also difficult to predict, meaning that a sweep across the cooperative ratio is needed for an effective implementation of the cooperative method.

\subsection{Dynamic cooperative ratios}

\begin{figure}
    \centering
    \includegraphics[width = \linewidth]{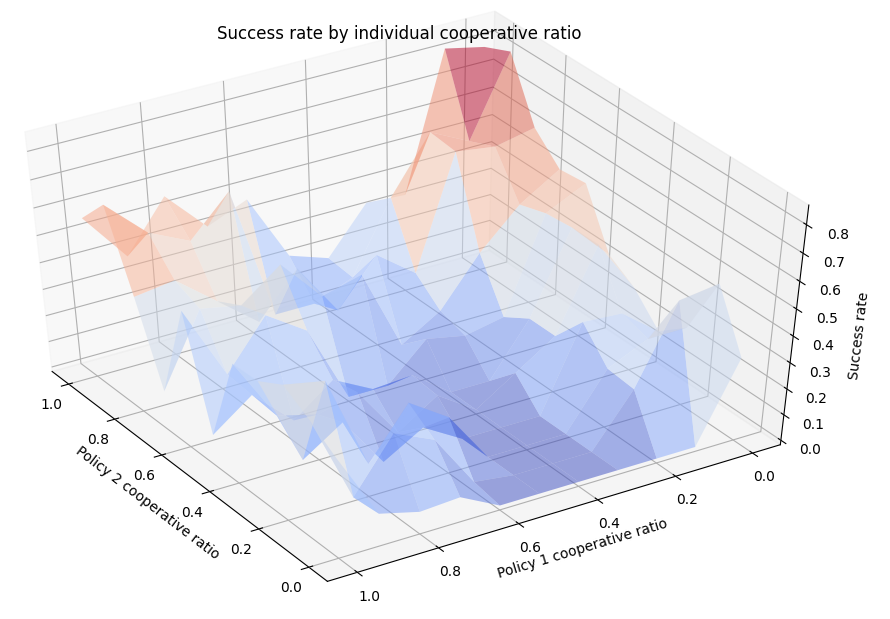}
    \caption{The measured success rate in the 3 room maze domain with different cooperative ratios for each cooperative policy. The success rate is an average success rate across the last 10 epochs of each of the 10 iterations of each configuration. Blue represents a low success rate and red represents a high success rate.}
    \label{fig:indCRatios}
\end{figure}

The results of the individual cooperative ratio experiments are shown in Figure \ref{fig:indCRatios}. Using the individually tuned cooperative ratios, a higher level of performance was found in the 3 room domain. Using cooperative ratios of 0.1 and 1.0 for the first two policies respectively result in the highest level of success, far outperforming the policies trained using a shared cooperative ratio. Though these results are higher than the shared cooperative ratio results, each additional cooperative policy is an additional parameter, which can quickly become a very expensive number of variables to sweep over. 

\begin{figure}
    \centering
    \includegraphics[width=\linewidth]{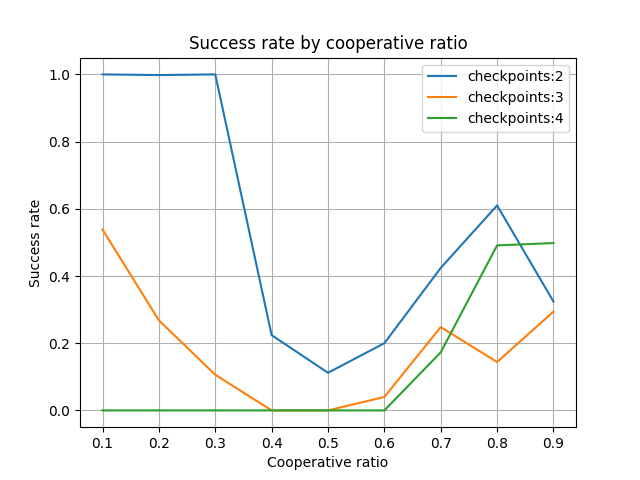}
    \caption{The measured success rate in different domains using different cooperative ratios. The success rate is an average success rate across the last 50 epochs of each of the 10 iterations of each configuration.}
    \label{fig:successByCRatio}
\end{figure}

Some insights can be gathered from analysing the effect on success rate when changing the cooperative ratio by looking at Figures \ref{fig:successByCRatio} and \ref{fig:indCRatios}. These figures are domain-specific, and as a result, the findings are also most likely domain-specific. In all of the shared ratio experiments, using a cooperative ratio around 0.5 produces a low level of performance. This is likely due to the two different motives of a cooperative policy; maximising your current reward and enabling the next policy to maximise reward, are at odds enough that comparing them on an equivalent scale is quite confusing for the policy to learn from, and this leads to poor performance. This trend can be seen in the individual cooperative ratio experiments when comparing the cooperative ratios for the first policy. Using a cooperative ratio of 0.5 for the first policy leads to lower performance similar to that seen in the shared ratio experiments. 

The individual ratio experiments also show that using a low cooperative ratio for the first policy has a higher level of performance than using a high cooperative ratio, a finding that is mirrored in the 2 room results in the shared ratio experiments. This shows that in the first subtask of this domain it is more difficult to avoid the dead-end without increased direction from the subsequent critic. This is likely due to the extra complexity introduced from the random initialisation. 

\section{Conclusion}

The CSAC algorithm is more effective in the maze domain than the standard RL approach, either by using a single agent or by splitting the domain into uncooperative policies. In the simplest domain (2 room maze; Figure \ref{fig:2RoomResults}), CSAC converged on a solution 4 times faster than the single agent and was able to maintain a high level of performance that the uncooperative policies were not able to maintain. In the more complex domains (3 and 4 room mazes; Figures \ref{fig:3RoomResults}, \ref{fig:4RoomResults}), the cooperative policies had a consistently higher level of performance than the uncooperative policies, whereas the single agent was not able to find any solution to the task within 3 million training steps. This shows that using the cooperative policies method can lead to consistently improved performance than using either of the other two conventional methods.

The proposed CSAC algorithm has some limitations. Firstly, CSAC requires that the task be decomposed into subtasks. There are works that seek to learn the decomposition of a task \cite{Andreas2017}, and these methods may be able to be combined with for a more autonomous algorithm. Secondly, a reward signal needs to defined for each subtask. The field of Inverse Reinforcement Learning seeks to autonomously learn a reward signal for a given task \cite{fu2017learning}, and using the advances in that field this reward definition requirement may be overcome. Thirdly, the cooperative ratio variable is required to be defined for each subtask. A potential solution to this issue is to utilise meta-learning \cite{frans2017meta} to tune this parameter. Finally, the CSAC method requires that the overall task be sequential. The CSAC algorithm improved performance comes from the fact that it utilises this assumption, but it is a limitation as many tasks can't be defined as sequential.

The CSAC algorithm would be extended well by finding a way to automatically tune the cooperative ratio \(\eta\). Currently, the value needs to be found through a sweep of an environment, as there is currently no algorithm developed to tune it. Developing a way to tune the cooperative ratio would allow for a far more efficient training than utilising a sweep.

The maze domain that is used in this paper is relatively simple. CSAC is capable of extending to more complex domains, and it is expected that more difficult domains would further highlight the effectiveness of this method. The benefits of splitting domains are more beneficial in larger domains, and so testing this method in higher dimensional domains should produce similarly effective results.

\newpage
\bibliographystyle{IEEEtran}
\bibliography{references.bib}


\end{document}